\documentclass{article}
\usepackage{spconf,amsmath,graphicx,hyperref,amssymb,amsfonts}
\usepackage[a4paper, margin=1in]{geometry} 
\usepackage{booktabs} 
\usepackage{multirow}
\usepackage{pgfplots}
\pgfplotsset{compat=1.18}
\usepackage[font=small]{caption}
\usepackage[font=footnotesize]{subcaption}

\title{Jointly Conditioned Diffusion Model for Multi-View Pose-Guided Person Image Synthesis}

\name{Zhi Gong$^{1}$\textsuperscript{$\dagger$}\thanks{\textsuperscript{$\dagger$}\ Equal contribution. \textsuperscript{*}\ Corresponding authors.}, 
      Chengyu Xie$^{1}$\textsuperscript{$\dagger$}, 
      Junchi Ren$^{1}$,
       Linkun Yu$^{1}$, 
      Si Shen$^{1}$, 
      Fei Shen$^{2}$\textsuperscript{*},
      Xiaoyu Du$^{1}$,
      }
\address{$^{1}$ Nanjing University of Science and Technology, China\\
         $^{2}$ NExT++ Research Centre, National University of Singapore, Singapore}

\begin{document}

\maketitle

\begin{abstract}
Pose-guided human image generation is limited by incomplete textures from single reference views and the absence of explicit cross-view interaction. We present jointly conditioned diffusion model (JCDM), a jointly conditioned diffusion framework that exploits multi-view priors. The appearance prior module (APM) infers a holistic identity preserving prior from incomplete references, and the joint conditional injection (JCI) mechanism fuses multi-view cues and injects shared conditioning into the denoising backbone to align identity, color, and texture across poses. JCDM supports a variable number of reference views and integrates with standard diffusion backbones with minimal and targeted architectural modifications. Experiments demonstrate state of the art fidelity and cross-view consistency. The code is available at \href{https://github.com/DusanYule1/JCDM}{\texttt{DusanYule1/JCDM}}.
\end{abstract}


\section{Introduction}
\label{sec:intro}
Diffusion models~\cite{bhunia2023person,han2023controllable,shen2023advancing} have driven rapid progress in controllable image generation, especially for human centric human image synthesis. Pose-guided person image synthesis aims to generate high fidelity images that preserve a person's identity while following a target pose, given one or more reference images. Typical applications include virtual try-on, digital avatar creation, and content production.

Recent state-of-the-art methods enhance appearance transfer and pose alignment via tailored architectures, following two main directions. (1) Appearance feature injection: PIDM~\cite{bhunia2023person} diffuses textures in pixel space, while PoCoLD~\cite{han2023controllable} applies latent-space diffusion with DensePose~\cite{guler2018densepose} constraints. GRPose~\cite{yin2025grpose} models body-part relations as a graph to better preserve structure, and MCLD~\cite{liu2025multi} introduces multi-focal conditions on faces and clothing to reduce detail loss from latent compression. (2) Semantic and pose guidance: CFLD~\cite{lu2024coarse} learns coarse semantic prompts to mitigate overfitting, and Stable-Pose~\cite{wang2024stable} uses ViTs with coarse-to-fine attention masking to improve pose alignment, especially for complex poses. To bridge large pose gaps, PCDMs~\cite{shen2023advancing} adopt staged progressive generation, while IMAGPose~\cite{shen2024imagpose} unifies multi-view inputs. Nonetheless, relying on a single reference limits texture coverage and causes identity drift, and the absence of explicit cross-view interaction during parallel synthesis leads to appearance inconsistencies.

Despite these advances, practical deployment remains hindered by three factors. First, single-view conditioning provides incomplete coverage of self occluded and fine textured regions, which degrades identity preservation and causes artifacts under large pose changes. Second, most parallel pipelines lack explicit cross-view interaction, so appearance cues are not reconciled across poses, yielding inconsistencies in color, texture, and local details. Third, single image inference dominates current designs, which increases latency and computational cost when multiple reference views are available; in real applications, multi-view inputs are common and should be exploited within one coherent generation process.

\begin{figure*}[t]
  \centering
  \includegraphics[width=0.95\linewidth]{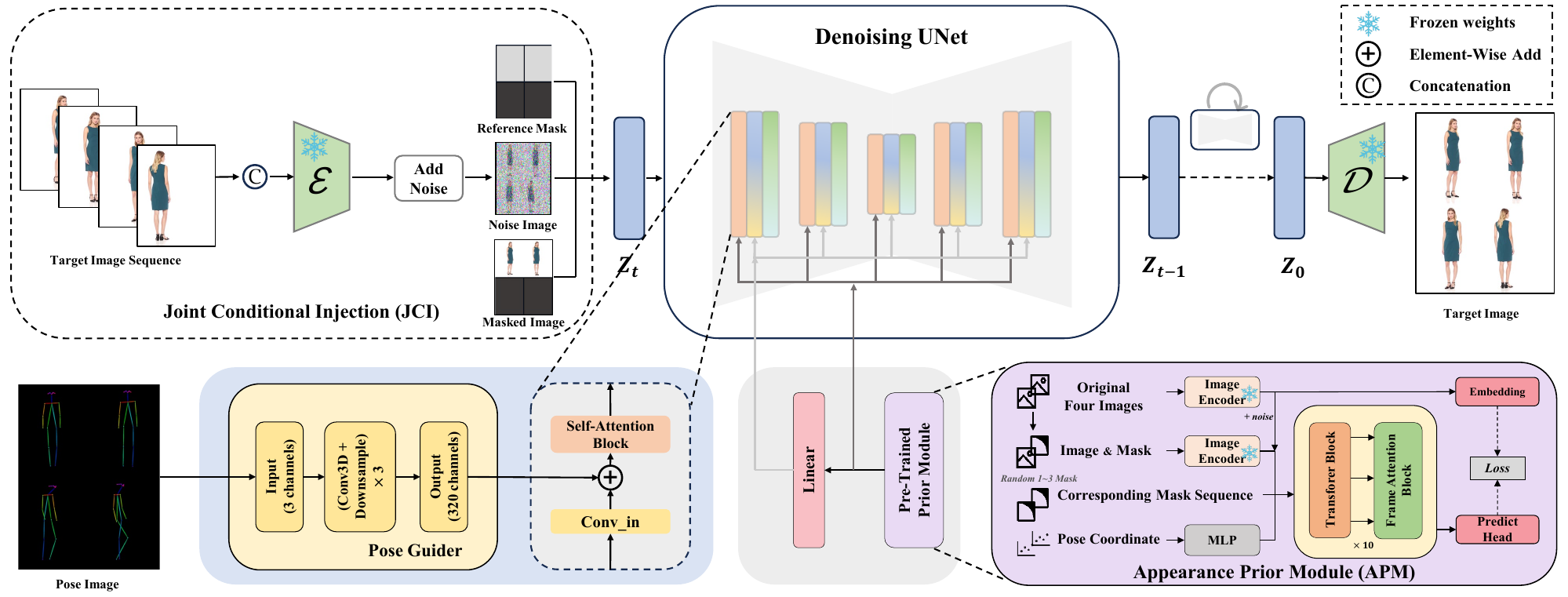}
  \vspace{-0.3cm}
\caption{
Overview of proposed JCDM. JCI encodes multiple reference views and target poses into a unified nine channel latent for the denoising UNet. APM infers an identity preserving appearance prior from the same inputs. The prior and the latent jointly condition the UNet to render consistent high fidelity images across poses.
}
  \label{fig:architecture}
    \vspace{-0.6cm}
\end{figure*}

To address these challenges, we propose Jointly Conditioned Diffusion Model (JCDM), a unified framework that exploits multi-view priors for consistent, faithful synthesis. JCDM comprises two plug-and-play modules compatible with existing diffusion backbones: Appearance Prior Module (APM) predicts a holistic identity-preserving prior from incomplete references to recover missing textures, and Joint Conditional Injection (JCI) fuses multi-view attributes and injects them as shared conditioning, aligning identity, color, and texture across poses. The design naturally supports a variable number of reference views at inference.
Our main contributions are as follows:
\begin{itemize}
    \item We propose JCDM, a unified jointly conditioned diffusion framework that exploits multi-view priors for pose-guided person image synthesis.
    \item JCDM comprises two components: an APM inferring a holistic identity-preserving prior from incomplete references, and a JCI mechanism aligning multi-view attributes.
    \item JCDM achieves state of the art fidelity and cross-view appearance alignment with lower multi-view inference latency on standard benchmarks.
\end{itemize}

\vspace{-0.2cm}
\section{Method}
\label{sec:method}

\subsection{Overview}
We propose the JCDM to address incomplete texture coverage and cross-view inconsistency in pose-guided generation. As shown in Fig.~\ref{fig:architecture}, JCDM comprises two components: an APM that infers a holistic identity preserving prior from sparse references (Section~\ref{sec:apm}), and a JCI mechanism that injects this prior together with multi-view cues into the denoising backbone (Section~\ref{sec:jci}). The APM ensures a consistent semantic representation, while the JCI mechanism effectively delivers and fuses it with conditional signals for the U-Net, enabling high-fidelity and cross-view consistent synthesis within a unified model. Finally, Training and inference strategies, including dual optimization objectives and classifier-free guidance, are detailed in Section~\ref{sec:training_inference} to complete the framework.

\subsection{Appearance Prior Module}\label{sec:apm}
\noindent\textbf{Motivation.} Existing methods struggle to infer plausible content for unseen regions when references are incomplete, for example, only a frontal view. The APM upgrades the task from appearance transfer to content inference by predicting a globally consistent semantic prior from limited visual cues. By learning a subject-specific prior in the CLIP space, APM imposes global constraints that stabilize identity and suppress hallucinated textures under large pose changes.

\noindent\textbf{Architecture.} The APM is a Transformer-based diffusion model that learns to reconstruct a complete semantic embedding from masked multi-view inputs. We adopt features from a frozen CLIP image encoder~\cite{radford2021learning} as image-level semantic representations. For each training sample, we form a sequence of $K$ images of the same subject under different poses. We randomly mask $k \in \{1,\dots,K-1\}$ views, and pass both the masked and the complete sequences through the frozen encoder to obtain the hidden visual representation $V_h$, the binary mask sequence $M$, and the target representation $V_t$. Pose coordinates from all views are encoded by a multilayer perceptron to yield pose features $P$. During training, noise is added to $V_t$ to obtain $V_n$. The input to the Transformer consists of the concatenation of $V_n$, $V_h$, $M$, $P$, and the timestep embedding $T_{\text{emb}}$. The APM predicts the denoised target embedding $\hat{V}_t$, thereby learning the mapping from sparse visual context and pose to a complete appearance prior. The masked and visible tokens are processed by stacked self attention and cross attention blocks where the noisy target tokens attend to visible view tokens and pose tokens. We add learnable view index embeddings and sinusoidal pose encodings to preserve ordering and geometry, and use a linear projection head to map outputs back to the CLIP embedding dimension. Training follows the standard diffusion denoising objective with a cosine noise schedule, and at inference the denoiser iteratively refines $V_n$ to recover $\hat{V}_t$ used as the appearance prior.

\subsection{Joint Conditional Injection}\label{sec:jci}
\noindent\textbf{Motivation.} When synthesizing multiple target poses in parallel, identity, clothing, and style must remain consistent across views. Prior methods seldom enable explicit cross-view interaction during generation. JCI enforces cross-view consistency by constructing a joint input and applying shared conditioning throughout the denoising process.

\noindent\textbf{Architecture.} JCI forms a joint conditional input with three components. First, all target views are spatially concatenated into a joint target image, which a frozen variational autoencoder (VAE) encodes~\cite{kingma2013auto} to a latent $x_0$ that is then perturbed to $x_t$. Second, a joint mask image places available reference views in fixed slots and masks unknown target slots; encoding this image yields a masked latent $z_m$. Third, a binary mask $M$ marks generation and reference regions. Concatenating these tensors along the channel dimension gives the UNet input, as follows,
\vspace{-.3cm}
\begin{equation}
% \vspace{-.3cm}
    z_{in} = \text{Concat}_{\text{channel}}(x_t, z_m, M).
\label{eq:jci_input}
\vspace{-.3cm}
\end{equation}
At every UNet block, the APM prior $V_{\text{pred}}$ is injected via cross attention as a shared condition. In addition, temporal attention treats the tiled views as a pseudo sequence and computes inter-view attention, enabling explicit feature level interaction during denoising. Here, we innovatively repurpose temporal attention to capture spatial consistency across views, rather than model temporal dynamics.

\subsection{Training and Inference}\label{sec:training_inference}
\noindent\textbf{Training.} JCDM uses two objectives. The APM is optimized to regress the clean semantic prior:
\vspace{-0.3cm}
{\small
\begin{equation}
\mathcal{L}_{\text{APM}}
= \mathbb{E}_{V_t, \epsilon, V_h, M, P, t}
\left\| V_t - \mathrm{APM}_{\theta}\!\left(V_n, V_h, M, P, t\right) \right\|_2^2 .
\label{eq:apm_loss}
\end{equation}
}
After training the APM, we freeze it and train the diffusion model to predict the noise on the joint target latent. Let $F_I$ and $F_P$ denote image and pose condition features extracted for JCI. The denoising objective is
\vspace{-0.3cm}
{\small
\begin{equation}
\mathcal{L}_{\text{JCDM}}
= \mathbb{E}_{x_0, \epsilon, V_{\text{pred}}, F_I, F_P, t}
\left\| \epsilon - \epsilon_{\theta}\!\left(x_t, V_{\text{pred}}, F_I, F_P, t\right) \right\|_2^2 ,
\label{eq:jcdm_loss}
\end{equation}
}
where $V_{\text{pred}}$ is the APM predicted appearance prior.

\noindent\textbf{Inference.} For a given subject, the APM is first run once to generate a fixed appearance prior, which is then held constant throughout the main diffusion process. We employ classifier free guidance to strengthen conditioning. The final noise prediction is
\vspace{-0.3cm}
{\small
\begin{equation}
\begin{aligned}
\hat{\epsilon}_{\theta}\!\left(x_t, V_{\text{pred}}, F_I, F_P, t\right)
= {}& w \, \epsilon_{\theta}\!\left(x_t, V_{\text{pred}}, F_I, F_P, t\right) \\
& + \left(1 - w\right) \, \epsilon_{\theta}\!\left(x_t, t\right),
\end{aligned}
\label{eq:cfg}
\vspace{-0.3cm}
\end{equation}
}
where $w$ is the guidance scale controlling the strength of the conditions.

\begin{table}[t]
\centering
\vspace{-.5cm}
\caption{Quantitative comparison on DeepFashion dataset.}
 \vspace{-0.3cm}
\resizebox{0.47\textwidth}{!}{
\label{tab:comparison}
\begin{tabular}{llccc}
\toprule
Dataset & Methods & SSIM ($\uparrow$) & LPIPS ($\downarrow$) & FID ($\downarrow$) \\
\midrule
\multirow{8}{*}{\begin{tabular}[c]{@{}c@{}}DeepFashion~\cite{liu2016deepfashion}\\ (256 $\times$ 176)\end{tabular}} 
& ADGAN\cite{men2020controllable} & 0.6721 & 0.2283 & 14.458 \\
& GFLA\cite{ren2020deep} & 0.7074 & 0.2341 & 10.573 \\
& DPTN\cite{zhang2022exploring} & 0.7112 & 0.1931 & 11.387 \\
& CASD\cite{zhou2022cross} & 0.7248 & 0.1936 & 11.373 \\
& NTED\cite{ren2022neural} & 0.7182 & 0.1752 & 8.6838 \\
& PIDM\cite{bhunia2023person} & 0.7312 & 0.1678 & 6.3671 \\
& \textbf{Ours} & \textbf{0.7619} & \textbf{0.1206} & \textbf{5.8325} \\
\midrule
\multirow{5}{*}{\begin{tabular}[c]{@{}c@{}}DeepFashion~\cite{liu2016deepfashion}\\ (512 $\times$ 352)\end{tabular}} 
& NTED\cite{ren2022neural} & 0.7376 & 0.1980 & 7.7821 \\
& PIDM\cite{bhunia2023person} & 0.7419 & 0.1768 & 5.8365 \\
& PoCoLD\cite{han2023controllable} & 0.7430 & 0.1920 & 8.4163 \\
& CFLD\cite{lu2024coarse} & 0.7478 & 0.1819 & 7.1490 \\
& \textbf{Ours} & \textbf{0.7827} & \textbf{0.1355} & \textbf{5.5347} \\
\bottomrule
\end{tabular}
}
\vspace{-.3cm}
\end{table}

\begin{figure}[t]
\vspace{-0.25cm}
    \centering
    \includegraphics[width=0.9\linewidth]{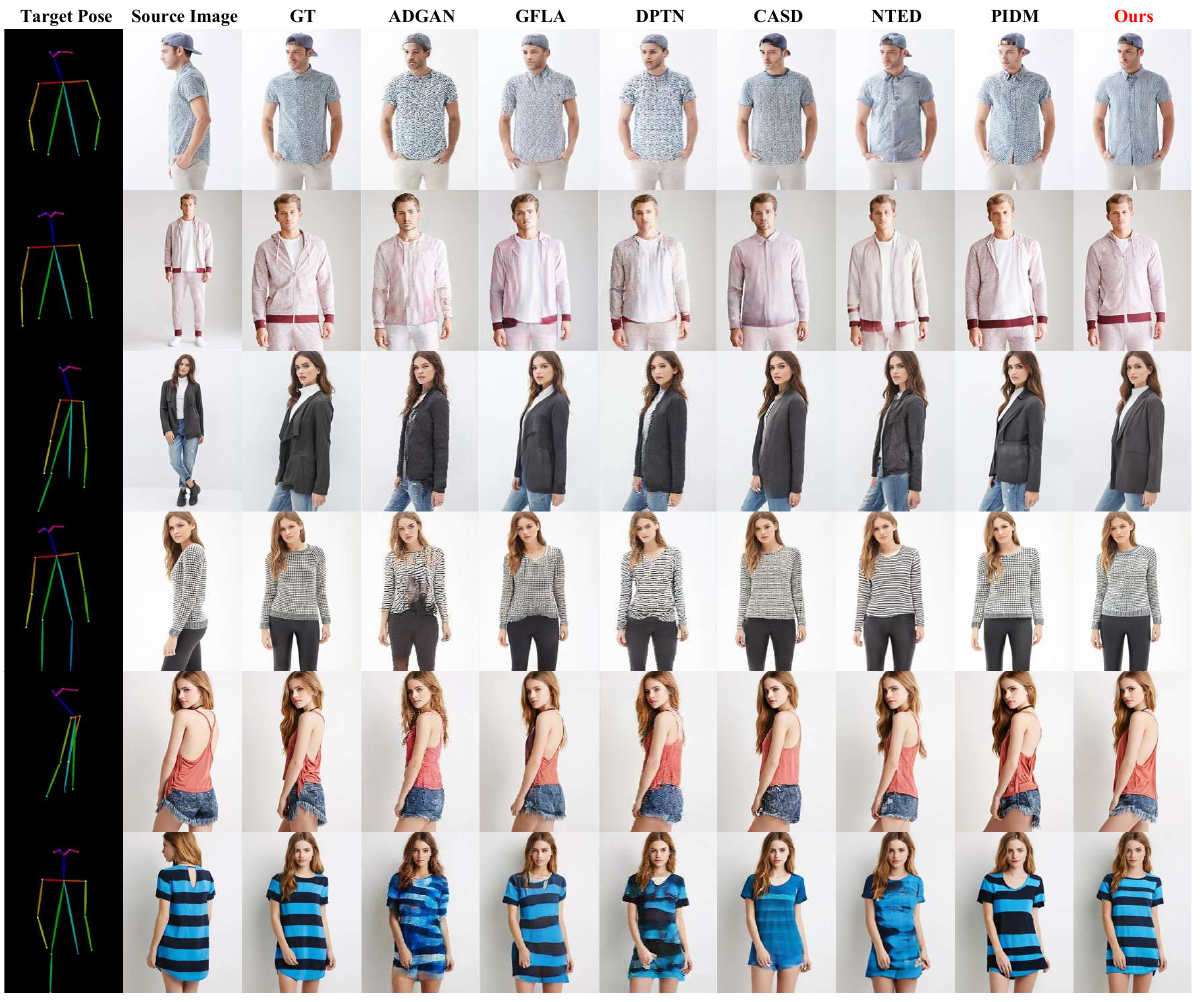}
    \vspace{-0.4cm}
    \caption{Qualitative comparison with SOTA methods.}
    \label{fig:qualitative_comparison}
      \vspace{-0.8cm}
\end{figure}

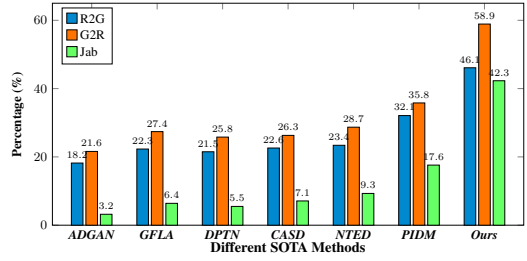
\begin{figure}[t]
\vspace{-0.3cm}
\centering
\resizebox{0.43\textwidth}{!}{%
    \begin{tikzpicture}
    \begin{axis}[
        ybar=0pt, bar width=8pt,
        width=13cm, height=7cm,
        ymin=0, ymax=65,
        ylabel={\textbf{Percentage (\%)}},
        ylabel style={font=\small},
        % --- Updated X-axis labels ---
        symbolic x coords={ADGAN,GFLA,DPTN,CASD,NTED,PIDM,Ours},
        xtick=data,
        x tick label style={font=\bfseries\itshape\small},
        yticklabel style={font=\small},
        tick align=inside,
        major grid style={dashed,gray!40},
        xlabel={\textbf{Different SOTA Methods}},
        xlabel style={font=\bfseries, yshift=5pt},
        legend style={
            at={(0.02,0.98)},
            anchor=north west,
            font=\small,
            cells={anchor=west},
        },
        legend image code/.code={%
            \draw[mark options={solid},fill=#1] 
            (0cm,-0.1cm) rectangle (0.25cm,0.15cm);
        },
        nodes near coords,
        nodes near coords align={vertical},
        every node near coord/.append style={font=\scriptsize, black},
        bar shift auto,
    ]
    
    \addplot+[fill=cyan!70!blue,draw=black] coordinates
     {(ADGAN,18.2) (GFLA,22.3) (DPTN,21.5) (CASD,22.6) (NTED,23.4) (PIDM,32.1) (Ours,46.1)};

    \addplot+[fill=orange!90!red,draw=black] coordinates
     {(ADGAN,21.6) (GFLA,27.4) (DPTN,25.8) (CASD,26.3) (NTED,28.7) (PIDM,35.8) (Ours,58.9)};

    \addplot+[fill=green!60!white,draw=black] coordinates
     {(ADGAN,3.2) (GFLA,6.4) (DPTN,5.5) (CASD,7.1) (NTED,9.3) (PIDM,17.6) (Ours,42.3)};
    
    \legend{R2G,G2R,Jab}
    
    \end{axis}
    \end{tikzpicture}
}
\vspace{-.3cm}
\caption{User study results.}
\label{fig:user_study}
\vspace{-.5cm}
\end{figure}

\section{Experiments}

\subsection{Implementation Details}
\label{sec:impl}

\noindent\textbf{Datasets.}
Following prior work~\cite{bhunia2023person,han2023controllable,shen2023advancing}, we evaluate on cleaned DeepFashion~\cite{liu2016deepfashion} (18{,}252 images) and an in-house video corpus (145,323 frames), which features subjects in diverse, uncontrolled environments with varied lighting and complex motions, providing a challenging testbed for model robustness. Poses are extracted with DWPose~\cite{yang2023effective} using body skeletons only, and train and test sets do not overlap. 

\begin{table}[t]
    \centering
    \caption{Ablation study on proposed JCDM.}
     \vspace{-0.3cm}
    \label{tab:ablation}
    \resizebox{0.47\textwidth}{!}{
    \begin{tabular}{l|ccc}
        \toprule
        Methods & SSIM ($\uparrow$) & LPIPS ($\downarrow$) & FID ($\downarrow$) \\
        \midrule
        B0 (Baseline)                      & 0.7215 & 0.1738 & 8.3562 \\
        B1 (B0 + JCI's Joint Image)        & 0.7390 & 0.1651 & 7.9134 \\
        B2 (B1 + JCI's Interaction)        & 0.7761 & 0.1423 & 6.8851 \\
        \textbf{JCDM (Ours)}     & \textbf{0.7827} & \textbf{0.1355} & \textbf{5.5347} \\
        \bottomrule
    \end{tabular}
    }
     \vspace{-0.4cm}
\end{table}

\noindent\textbf{Metrics.}
We evaluate with SSIM~\cite{wang2004image}, LPIPS~\cite{zhang2018unreasonable}, and Fr\'echet Inception Distance (FID)~\cite{heusel2017gans}. Human studies report real-to-generated (R2G)~\cite{ma2017pose}, generated-to-real (G2R)~\cite{ma2017pose}, and pairwise preference (JAB)~\cite{siarohin2018deformable}.

\noindent\textbf{Hyperparameters.}
Training uses eight NVIDIA A800 GPUs. APM has ten Transformer and frame attention blocks with width 2048. The diffusion backbone is Stable Diffusion v1.5, and the first UNet convolution is expanded to match the JCI input channels. Images are resized to \(768\times768\). We use AdamW with learning rate \(1\times10^{-5}\). Batch sizes are 32 for APM and 7 for JCDM. Both stages use a linear noise schedule with 1000 timesteps. Inference uses DDIM with 30 steps and classifier free guidance scale \(w=2.0\).

\subsection{Main Comparisons}
JCDM supports single and multi target synthesis. For fair comparison and to match the joint input format, we pad the masked layout with three black images and repeat the target pose three times at inference. The model outputs three identical target images, from which we randomly select one for quantitative and qualitative evaluation. Unless stated otherwise, this is the default setting. We compare JCDM with state of the art methods including ADGAN~\cite{men2020controllable}, GFLA~\cite{ren2020deep}, DPTN~\cite{zhang2022exploring}, NTED~\cite{ren2022neural}, CASD~\cite{zhou2022cross}, PoCoLD~\cite{han2023controllable}, PIDM~\cite{bhunia2023person}, and CFLD~\cite{lu2024coarse}.

\begin{figure}[t]
    \centering
    \resizebox{0.47\textwidth}{!}{
    \begin{tikzpicture}
        \begin{axis}[
            title={JCDM Prediction Accuracy vs. Training Steps},
            xlabel={Training Steps},
            ylabel={Average Cosine Similarity},
            xmin=0, xmax=100000,
            ymin=0.7, ymax=1.0,
            xtick={0, 20000, 40000, 60000, 80000, 100000},
            xticklabels={0, 20k, 40k, 60k, 80k, 100k},
            ytick={0.7, 0.75, 0.8, 0.85, 0.9, 0.95, 1.0},
            legend pos=south east,
            ymajorgrids=true,
            grid style=dashed,
            width=0.8\textwidth,
            height=6cm,
        ]
        
        \addplot[color=blue, mark=*, smooth]
        coordinates {
            (0, 0.72) (10000, 0.88) (20000, 0.935) (40000, 0.949) (60000, 0.961) (80000, 0.973) (100000, 0.981)
        };
        \addlegendentry{JCDM Accuracy}
        \end{axis}
    \end{tikzpicture}
    }
     \vspace{-0.4cm}
\caption{Learning curve of JCDM over 100k steps. Accuracy, measured by average cosine similarity, improves steadily and converges.}
    \label{fig:jcdm_curve}
     \vspace{-0.5cm}
\end{figure}
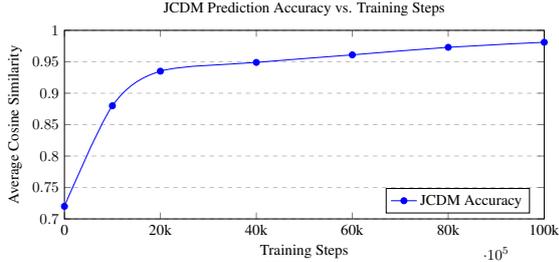

\begin{figure}[t]
    \centering
    \includegraphics[width=0.9\linewidth]{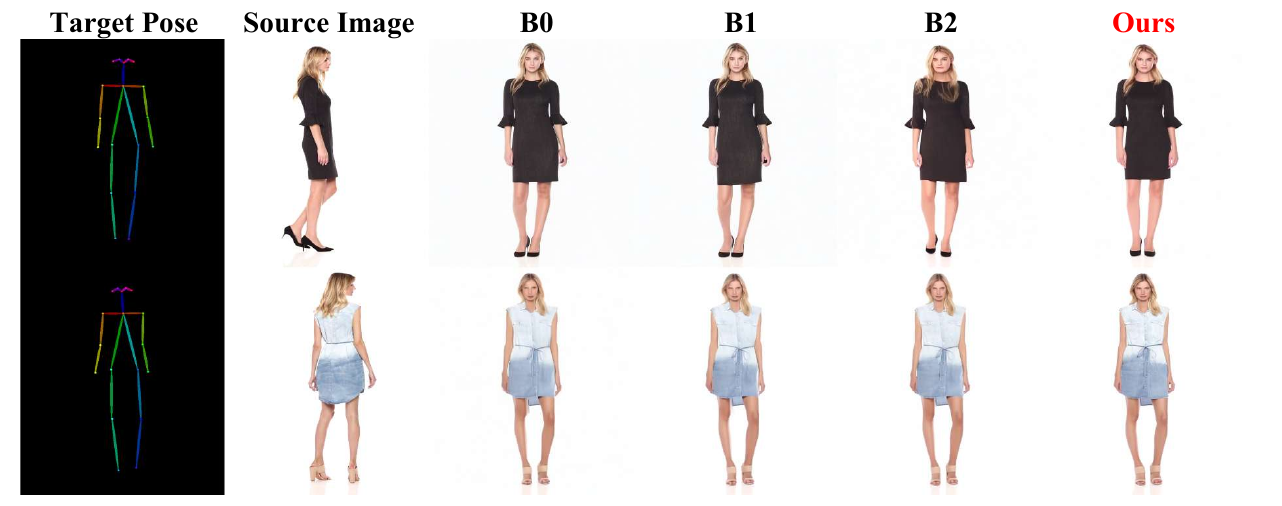}
    \vspace{-0.45cm}
    \caption{
        Ablation study results, showing the progressive visual improvements from baseline (B0) to JCDM (Ours).
    }
        \vspace{-0.3cm}
    \label{fig:qualitative_joint}
        \vspace{-0.3cm}
\end{figure}

\noindent\textbf{Quantitative Results.}
On Table~\ref{tab:comparison}, JCDM achieves the best SSIM/LPIPS/FID at both \(256\times176\) and \(512\times352\). At \(256\times176\), it surpasses GAN baselines (e.g., ADGAN~\cite{men2020controllable}), benefiting from JCI, which fuses multi-view cues and preserves source textures. At \(512\times352\), it outperforms diffusion based CFLD~\cite{lu2024coarse} due to APM’s stronger content priors. JCDM also exceeds PIDM~\cite{bhunia2023person} at both resolutions, validating the advantage of multi-view priors for high quality synthesis.

\noindent\textbf{Qualitative Results.}
Figure~\ref{fig:qualitative_comparison} shows that JCDM produces sharper details with fewer artifacts than diffusion baselines such as PIDM. For challenging views (e.g., synthesizing back views from frontal inputs), CNN methods like DPTN and NTED miss or distort textures, while JCDM infers complete, faithful content via the APM prior. JCI fuses multi-view cues to enable single pass multi pose synthesis with consistent identity and appearance. Overall, JCDM yields more realistic person images on DeepFashion.

\noindent\textbf{User Study.}
We conducted a study with 50 volunteers, evaluating real-to-generated (R2G), generated-to-real (G2R), and pairwise preference (JAB), where higher is better. As shown in Fig.~\ref{fig:user_study}, our method attains 58.5\% G2R, exceeding the second best by 18.4\%, and achieves 42.3\% on JAB, while also leading on R2G. These results align with the quantitative gains.

\vspace{-0.5cm}
\subsection{Ablation Study}
We ablate JCDM by adding modules stepwise. B0 uses single image CLIP features and a four channel UNet latent. B1 adds the JCI joint image input. B2 adds temporal attention for view interaction. The full model replaces CLIP features with the APM prior. Table~\ref{tab:ablation} and Fig.~\ref{fig:qualitative_joint} show consistent gains, with the largest from view interaction and the APM prior.

\noindent\textbf{Influence of the APM.}
Removing APM (B2) from the full model degrades FID and LPIPS (Table~\ref{tab:ablation}), showing that a learned appearance prior is essential for high fidelity synthesis. The APM learning curve (Fig.~\ref{fig:jcdm_curve}) exhibits steady gains in cosine similarity and stable convergence, indicating a reliable mapping from pose and sparse context to a global semantic appearance.

\begin{figure}[t]
    \centering
    \includegraphics[width=0.9\linewidth]{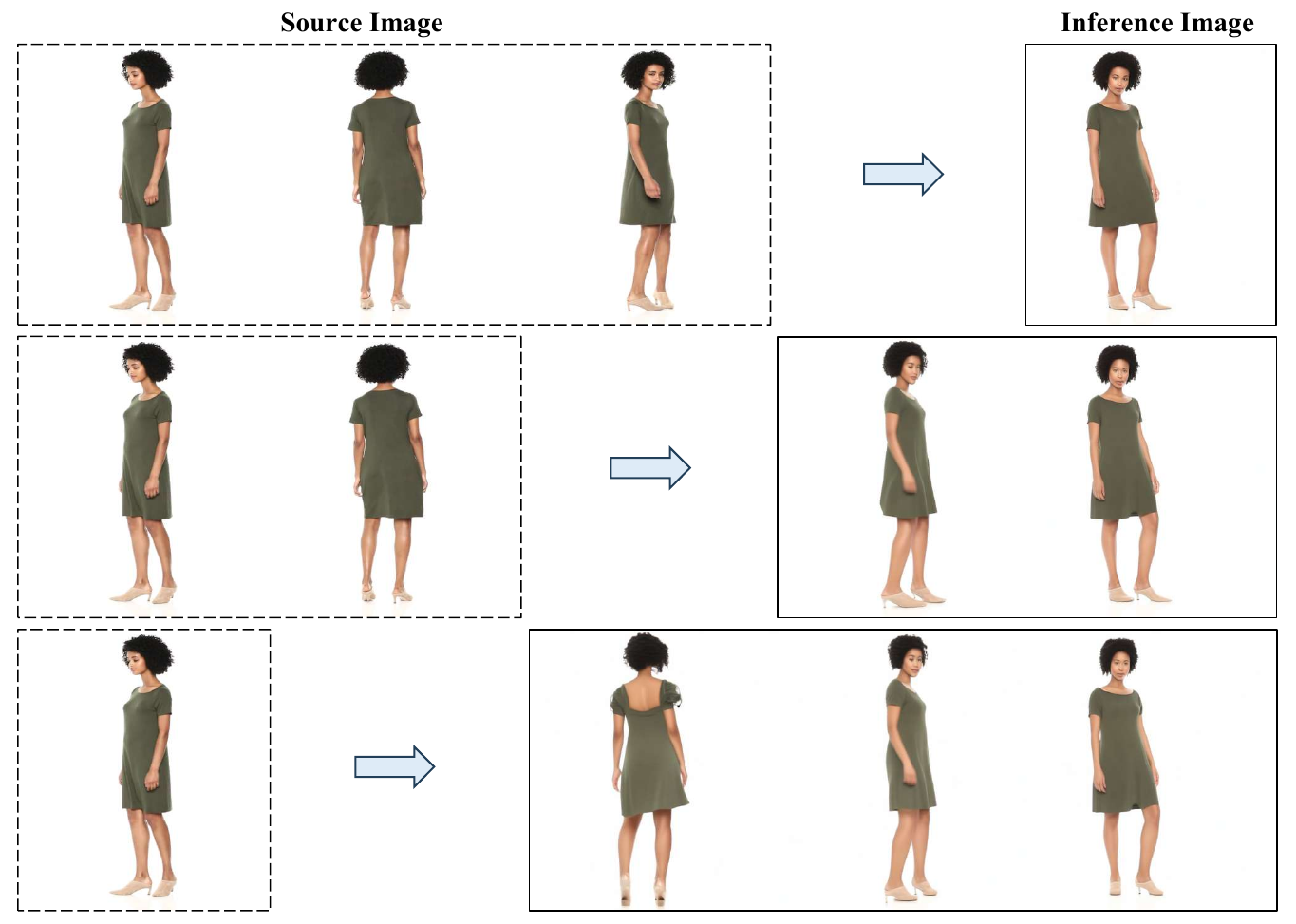}
    \vspace{-0.4cm}
    \caption{
        Results of multi-view generation capability.
    }
        \vspace{-0.4cm}
    \label{fig:multi_view_results}
        \vspace{-0.1cm}
\end{figure}

\begin{figure}[t]
    \centering
    \includegraphics[width=0.9\linewidth]{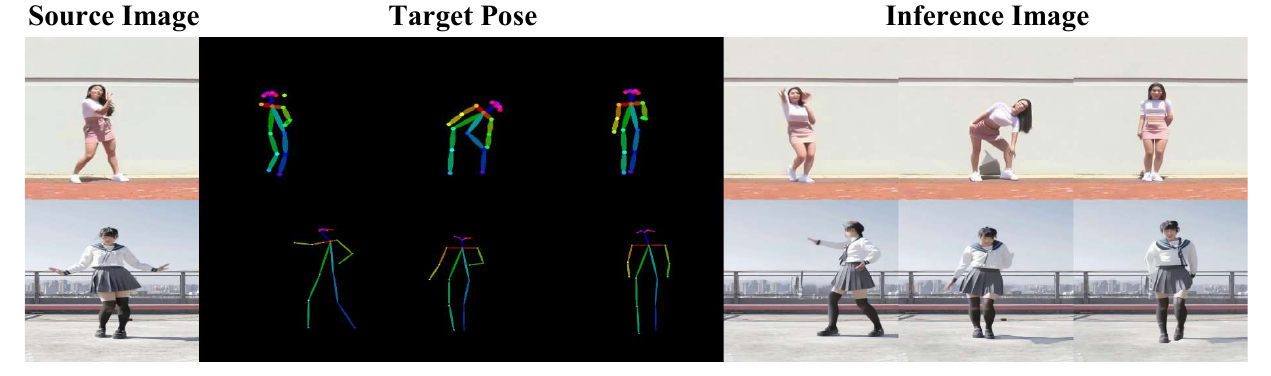}
       \vspace{-0.3cm}
    \caption{
        Generalization results on a challenging wild dataset.
    }
    \vspace{-0.3cm}
    \label{fig:Generalization}
       \vspace{-0.3cm}
\end{figure}

\noindent\textbf{Effectiveness of the JCI.}
Table~\ref{tab:ablation} shows steady gains from B0 to B2. Adding the joint image input (B1 vs B0) enables multi pose synthesis and improves all metrics. Enabling view interaction (B2 vs B1) yields the largest boost, with FID dropping from 7.9134 to 6.8851, confirming that explicit cross-view feature exchange is key for appearance consistency. Qualitative results in Fig.~\ref{fig:qualitative_joint} echo this trend, where residual clothing inconsistencies in B1 are corrected in B2.

\vspace{-0.5cm}
\subsection{More Results}

\noindent\textbf{Multi-View Generation.}
JCDM handles a variable number of reference and target views. Figure~\ref{fig:multi_view_results} shows results with one to three masks as conditions. The model adapts seamlessly and produces the same number of target images in one pass while preserving identity and appearance consistency. This single-pass approach significantly reduces inference latency compared to methods that require generating each view sequentially. This stems from the unified JCI design and supports practical content creation.

\noindent\textbf{Generalization.}
We evaluate on a wild dataset that differs from our static image data. As shown in Fig.~\ref{fig:Generalization}, JCDM produces high quality, identity consistent results on this out of domain distribution, demonstrating strong robustness and generalization.

\vspace{-0.5cm}
\section{Conclusion}
We addressed two key limitations in pose-guided person image synthesis: incomplete textures from single-view references and the absence of explicit cross-view interaction. We proposed JCDM, a jointly conditioned diffusion framework with two components, an Appearance Prior Module that infers a holistic identity preserving prior from sparse references, and a joint conditional injection mechanism that fuses multi-view cues as shared conditioning for the denoising backbone. The framework supports a variable number of reference views and integrates with standard diffusion backbones. Experiments demonstrate state of the art fidelity and cross-view consistency.

{\small
\bibliographystyle{IEEEbib}
\bibliography{strings,refs}
}
\end{document}